\documentclass[runningheads]{llncs}
\usepackage{graphicx}
\begin{document}
\title{Face Mask Detection using Transfer Learning of InceptionV3}
%
%
\author{G. Jignesh Chowdary\inst{1}\and
Narinder Singh Punn\inst{2}\and
Sanjay Kumar Sonbhadra\inst{2}\and
Sonali Agarwal\inst{2}}
\authorrunning{Chowdary, G.J. et al.}
%
\institute{Vellore Institute of Technology Chennai Campus\\
\email{guttajignesh.chowdary2018@vitstudent.ac.in}
\and
Indian Institute of Information Technology Allahabad\\
\email{\{pse2017002,rsi2017502,sonali\}@iiit.ac.in}}
\maketitle              
\begin{abstract}
The world is facing a huge health crisis due to the rapid transmission of coronavirus (COVID-19). Several guidelines were issued by the World Health Organization (WHO) for protection against the spread of coronavirus. According to WHO, the most effective preventive measure against COVID-19 is wearing a mask in public places and crowded areas. It is very difficult to monitor people manually in these areas. In this paper, a transfer learning model is proposed to automate the process of identifying the people who are not wearing mask. The proposed model is built by fine-tuning the pre-trained state-of-the-art deep learning model, InceptionV3. The proposed model is trained and tested on the Simulated Masked Face Dataset (SMFD). Image augmentation technique is adopted to address the limited availability of data for better training and testing of the model. The model outperformed the other recently proposed approaches by achieving an accuracy of 99.9\% during training and 100\% during testing.

\keywords{Transfer Learning  \and SMFD dataset \and Mask Detection \and InceptionV3 \and Image augmentation.}
\end{abstract}
\section{Introduction}
In view of the transmission of coronavirus disease (COVID-19), it was advised by the World Health Organization (WHO) to various countries to ensure that their citizens are wearing masks in public places. Prior to COVID-19, only a few people used to wear masks for the protection of their health from air pollution, and health professionals used to wear masks while they were practicing at hospitals. With the rapid transmission of COVID-19, the WHO has declared it as the global pandemic. According to WHO \cite{1}, the infected cases around the world are close to 22 million. The majority of the positive cases are found in crowded and over-crowded areas. Therefore, it was prescribed by the scientists that wearing a mask in public places can prevent transmission of the disease \cite{2}. An initiative was started by the French government to identify passengers who are not wearing masks in the metro station. For this initiative, an AI software was built and integrated with security cameras in Paris metro stations \cite{3}.

Artificial intelligence (AI) techniques like machine learning (ML) and deep learning (DL) can be used in many ways for preventing the transmission of COVID-19 \cite{4}. Machine learning and deep learning techniques allow to forecast the spread of COVID-19 and helpful to design an early prediction system that can aid in monitoring the further spread of disease. For the early prediction and diagnosis of complex diseases, emerging technologies like the Internet of Things (IoT), AI, big data, DL and ML are being used for the faster diagnosis of COVID-19  \cite{5,6,7,8,9}.

The main aim of this work is to develop a deep learning model for the detection of persons who are not wearing a face mask. The proposed model uses the transfer learning of InceptionV3 to identify persons who are not wearing a mask in public places by integrating it with surveillance cameras. Image augmentation techniques are used to increase the diversity of the training data for enhancing the performance of the proposed model. The rest of the paper is divided into 5 sections and is as follows. Sections 2 deals with the review of related works in the past. Section 3 describes the dataset. Section 4 describes the proposed model and Section 5 presents the experimental analysis of the proposed transfer-learning model. Finally, the conclusion of the proposed work is presented in Section 6.

\section{Literature Review}
Over the period there have been many advancements in the deep learning towards object detection and recognition in various application domains~\cite{zhao2019object,punn2019crowd}. In general, most of the works focus on image reconstruction and face recognition for identity verification. But the main aim of this work is to identify people who are not wearing masks in public places to control the further transmission of COVID-19. Bosheng Qin and Dongxiao Li \cite{10} have designed a face mask identification method using the SRCNet classification network and achieved an accuracy of 98.7\% in classifying the images into three categories namely “correct facemask wearing”, “incorrect facemask wearing” and “no facemask wearing”. Md. Sabbir Ejaz et al. \cite{11} implemented the Principal Component Analysis (PCA) algorithm for masked and un-masked facial recognition. It was noticed that PCA is efficient in recognizing faces without a mask with an accuracy of 96.25\% but its accuracy is decreased to 68.75\% in identifying faces with a mask. In a similar facial recognition application, Park et al. \cite{12} proposed a method for the removal of sunglasses from the human frontal facial image and reconstruction of the removed region using recursive error compensation. 

Li et al. \cite{13} used YOLOv3 for face detection, which is based on deep learning network architecture named darknet-19, where WIDER FACE and Celebi databases were used for training, and later the evaluation was done using the FDDB database. This model achieved an accuracy of 93.9\%. In a similar research, Nizam et al. \cite{14} proposed a GAN based network architecture for the removal of the face mask and the reconstruction of the region covered by the mask. Rodriguez et al. \cite{15} proposed a system for the automatic detection of the presence or absence of the mandatory surgical mask in operating rooms. The objective of this system is to trigger alarms when a staff is not wearing a mask. This system achieved an accuracy of 95\%. Javed et al. \cite{16} developed an interactive model named MRGAN that removes objects like microphones in the facial images and reconstructs the removed region's using a generative adversarial network. Hussain and Balushi \cite{17} used VGG16 architecture for the recognition and classification of facial emotions. Their VGG16 model is trained on the KDEF database and achieved an accuracy of 88\%. 

Following from the above context it is evident that specially for mask detection very limited number of research articles have been reported till date whereas further improvement is desired on existing methods. Therefore, to contribute in the further improvements of face mask recognition in combat against COVID-19, a transfer learning based approach is proposed that utilizes trained inceptionV3 model.

\section{Dataset Description}
In the present research article, a Simulated Masked Face Dataset (SMFD) \cite{18} is used that consists of 1570 images that consists of 785 simulated masked facial images and 785 unmasked facial images. From this dataset, 1099 images of both categories are used for training and the remaining 470 images are used for testing the model. Few sample images from the dataset are shown in Fig. 1.
\begin{figure}
\includegraphics[width=\textwidth]{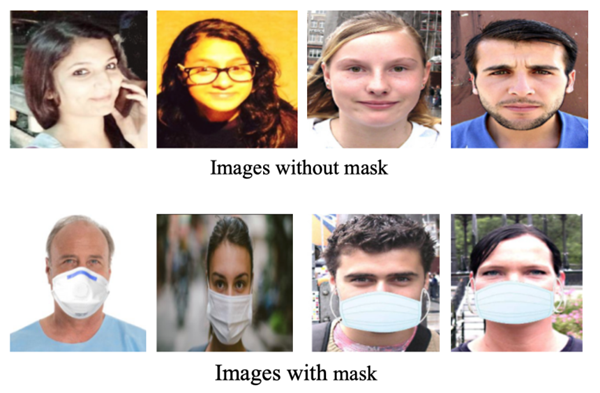}
\caption{Sample images from the dataset.} \label{fig1}
\end{figure}

\section{Proposed Methodology}
It is evident from the dataset description that there are a limited number of samples due to the government norms concerning security and privacy of the individuals. Whereas deep learning models struggle to learn in presence of a limited number of samples. Hence, over-sampling can be the key to address the challenge of limited data availability. Thereby the proposed methodology is split into two phases. The first phase deals with over-sampling with image augmentation of the training data whereas the second phase deals with the detection of face mask using transfer learning of InceptionV3. 

\subsection{Image augmentation}
Image augmentation is a technique used to increase the size of the training dataset by artificially modifying images in the dataset. In this research, the training images are augmented with eight distinct operations namely shearing, contrasting, flipping horizontally, rotating, zooming, blurring. The generated dataset is then rescaled to 224 x 224 pixels, and converted to a single channel greyscale representation. Fig. 2 shows an example of an image that is augmented by these eight methods.

\begin{figure}
\includegraphics[width=\textwidth]{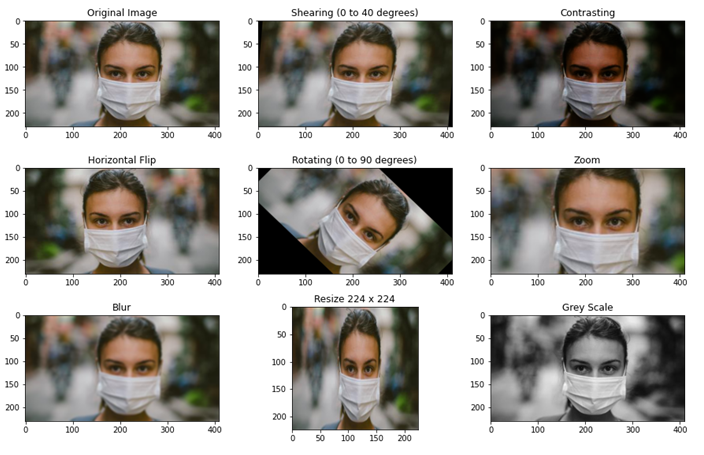}
\caption{Augmentation of training images.} \label{fig1}
\end{figure}

\subsection{Transfer Learning}
Deep neural networks are used for image classification because of their better performance than other algorithms. But training a deep neural network is expensive because it requires high computational power and other resources, and it is time-consuming. In order to make the network to train faster and cost-effective, deep learning-based transfer learning is evolved. Transfer learning allows to transfer the trained knowledge of the neural network in terms of parametric weights to the new model. Transfer learning boosts the performance of the new model even when it is trained on a small dataset. There are several pre-trained models like InceptionV3, Xception, MobileNet, MobileNetV2, VGG16, ResNet50, etc. \cite{19,20,21,22,23,24} that are trained with 14 million images from the ImageNet dataset. InceptionV3 is a 48 layered convolutional neural network architecture developed by Google.

In this paper, a transfer learning based approach is proposed that utilizes the InceptionV3 pre-trained model for classifying the people who are not wearing face mask. For this work, the last layer of the InceptionV3 is removed and is fine-tuned by adding 5 more layers to the network. The 5 layers that are added are an average pooling layer with a pool size equal to 5 x 5, a flattening layer, followed by a dense layer of 128 neurons with ReLU activation function and dropout rate of 0.5, and finally a decisive dense layer with two neurons and softmax activation function is added to classify whether a person is wearing mask. This transfer learning model is trained for 80 epochs with each epoch having 42 steps.  The schematic representation of the proposed methodology is shown in Fig. 3. The architecture of the proposed model is shown in Fig. 4.

\begin{figure}
\includegraphics[width=\textwidth]{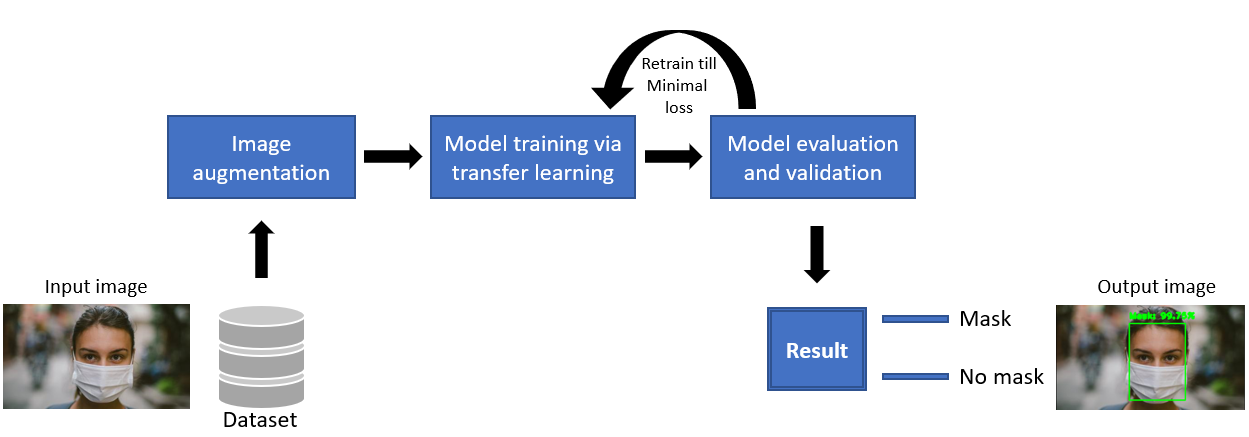}
\caption{Schematic representation of the proposed work.} \label{fig1}
\end{figure}

\begin{figure}
\includegraphics[width=\textwidth]{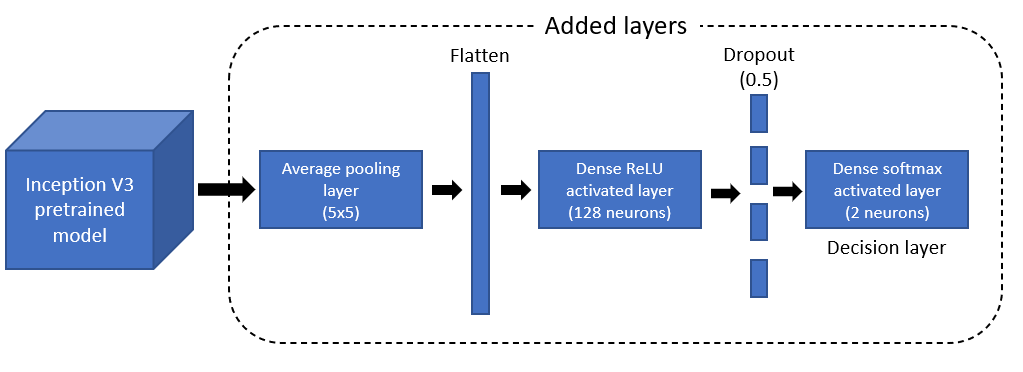}
\caption{Architecture of the proposed model.} \label{fig1}
\end{figure}

\section{Results}
The experimental trials for this work are conducted using the Google Colab environment. For evaluating the performance of the transfer learning model several performance metrics are used namely Accuracy, Precision, Sensitivity, Specificity, Intersection over Union (IoU), and Matthews Correlation Coefficient (MCC as represented as follows:

\begin{equation}
Accuracy=\frac{(TP+TN)}{(TP+TN+FP+FN)}               
\end{equation}

\begin{equation}
    Precision=\frac{TP}{(TP+FP)}
\end{equation}

\begin{equation}
    Sensitivity=\frac{TP}{(TP+FN)}
\end{equation}

\begin{equation}
    Specificity=\frac{TN}{(TN+FP)}
\end{equation}

\begin{equation}
    IoU=\frac{TP}{(TP+FP+FN)}
\end{equation}

\begin{equation}
    MCC = \frac{(TP\times TN)-(FP\times FN)}{\sqrt{(TP+FP)(TP+FN)(TN+FP)(TN+FN)}}
\end{equation}

\begin{equation}
    CE=\sum _{i=1}^{n}{Y}_{i}{\log}_{2}({p}_{i})
\end{equation}
In equation 7, Classification error is formulated in terms of Y\textsubscript{i} and p\textsubscript{i} where Y\textsubscript{i} represents the one-hot encoded vector and p\textsubscript{i} represents the predicted probablitity. The other performance metrics are formulated in terms of True Positive  (TP), False Positive (FP), True Negative (TN), and False Negative (FN). The TP, FP, TN and FN are represented in a grid-like structure called the confusion matrix. For this work, two confusion matrices are constructed for evaluating the performance of the model during training and testing. The two confusion metrics are shown in Fig. 5 whereas Fig. 6, shows the comparison of the area under the ROC curve, precision, loss and accuracy during training and testing the model.

The performance of the model during training and testing are shown in Table 1. The accuracy of the transfer-learning model is compared with other machine learning and deep learning models namely decision tree, support vector machine\cite{25} MobileNet, MobileNetV2, Xception, VGG16 and VGG19 when trained under the same environment, the proposed model achieved higher accuracy than the other models as represented in Fig. 7.

\begin{figure}
\includegraphics[width=\textwidth]{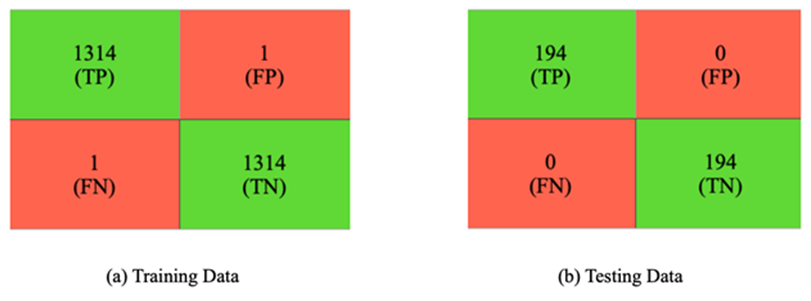}
\caption{Confusion Matrix.} \label{fig1}
\end{figure}

\begin{figure}
\includegraphics[width=\textwidth]{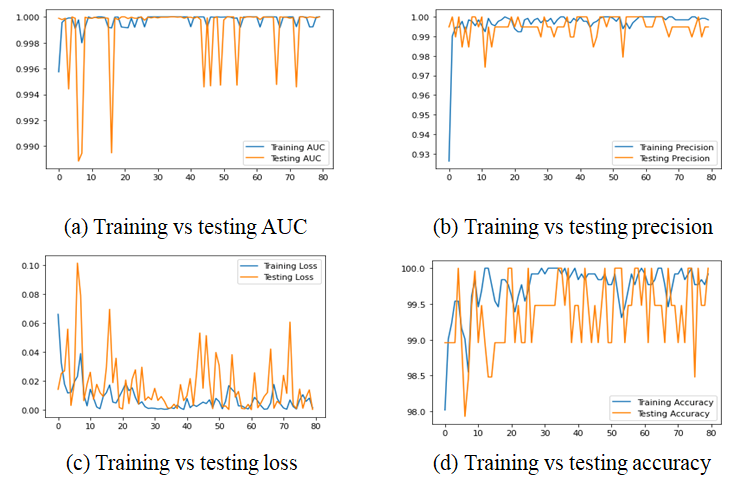}
\caption{Comparison of performance of the proposed model during training and testing.} \label{fig1}
\end{figure}

\begin{figure}
\includegraphics[width=\textwidth]{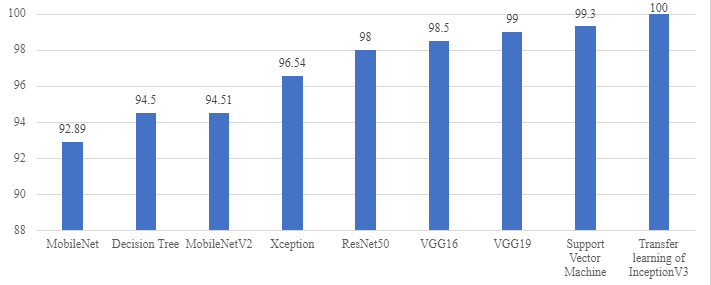}
\caption{Comparision of performance of the proposed model with other models.} \label{fig1}
\end{figure}

\begin{center}

\begin{table}
\centering
\caption{Performance of the proposed model.}\label{tab1}
\begin{tabular}{|l|l|l|}
\hline
\textbf{Performance Metrics} &  \textbf{Training}  & \textbf{Testing} \\
\hline

Accuracy &  99.92\% & 100\%\\

Precision &  99.9\% &	100\%\\

Specificity & 99.9\%	& 100\%\\

Intersection over Union (IoU) & 99.9\% &	100\%\\

Mathews Correlation Coefficient (MCC) & 0.9984	& 1.0\\

Classification Loss & 0.0015	& 3.8168E-04\\
\hline
\end{tabular}
\end{table}
\end{center}

\subsection{Output}
The main objective of this research is to develop an automated approach to help the government officials of various countries to monitor their citizens that they are wearing masks at public places. As a result, an automated system is developed that uses the transfer learning of InceptionV3 to classify people who are not wearing masks. For some random sample the output of the model shows the bounding box around the face where green and red color indicates a person is wearing mask and not wearing mask respectively along with the confidence score, as shown in Fig. 8.

\begin{figure}
\includegraphics[width=\textwidth]{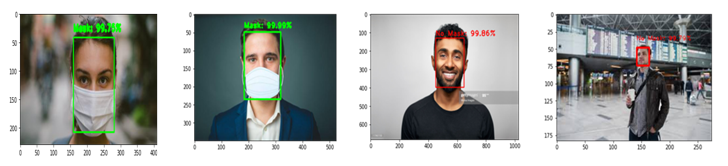}
\caption{Output from the proposed model.} \label{fig1}
\end{figure}

\section{Conclusion}
The world is facing a huge health crisis because of pandemic COVID-19. The governments of various countries around the world are struggling to control the transmission of the coronavirus. According to the COVID-19 statistics published by many countries, it was noted that the transmission of the virus is more in crowded areas. Many research studies have proved that wearing a mask in public places will reduce the transmission rate of the virus. Therefore, the governments of various countries have made it mandatory to wear masks in public places and crowded areas. It is very difficult to monitor crowds at these places. So in this paper, we propose a deep learning model that detects persons who are not wearing a mask. This proposed deep learning model is built using transfer learning of InceptionV3. In this work image augmentation techniques are used to enhance the performance of the model as they increase the diversity of the training data. The proposed transfer learning model achieved accuracy and specificity of 99.92\%, 99.9\% during training, and 100\%, 100\% during testing on the SMFD dataset. The same work can further be improved by employing large volumes of data and can also be extended to classify the type of mask, and implement a facial recognition system, deployed at various workplaces to support person identification while wearing the mask.

\section*{Acknowledgement}
This research is supported by “ASEAN- India Science \& Technology Development Fund (AISTDF)”, SERB, Sanction letter no. – IMRC/AISTDF/R\&D/P-6/2017. Authors are also thankful to the authorities of "Vellore Institute of Technology", Chennai, India and “Indian Institute of Information Technology Allahabad”, Prayagraj, India, for providing the infrastructure and necessary support.

%
%
%

 \bibliographystyle{splncs04}
 \bibliography{bibliographty.bib}

\end{document}